\documentclass{article}
\usepackage{spconf,amsmath,graphicx,multirow}

\usepackage{amsmath,graphicx}
\usepackage{amssymb,amsmath,epsfig}
\usepackage{lipsum,color,multirow,url}
\usepackage{arydshln, algorithm, algpseudocode}
\usepackage{microtype}
\usepackage{nth}
\usepackage[inline]{enumitem}
\usepackage{romannum}
\usepackage{xcolor}

\title{Streaming Attention-Based Models with Augmented Memory \\ for End-to-End Speech Recognition}
\name{
\begin{tabular}{c}
Ching-Feng Yeh, Yongqiang Wang, Yangyang Shi, Chunyang Wu,  \\
Frank Zhang, Julian Chan, Michael L. Seltzer
\end{tabular}
}

\address{Facebook AI, USA}

\begin{document}
%
\maketitle
\begin{abstract}
Attention-based models have been gaining popularity recently for their strong performance demonstrated in fields such as machine translation \cite{vaswani2017attention} and automatic speech recognition \cite{gulati2020conformer}.
One major challenge of attention-based models is the need of access to the full sequence and the quadratically growing computational cost concerning the sequence length.
These characteristics pose challenges, especially for low-latency scenarios, where the system is often required to be streaming.
In this paper, we build a compact and streaming speech recognition system on top of the end-to-end neural transducer architecture \cite{graves2012sequence} with attention-based modules augmented with convolution \cite{gulati2020conformer}.  
The proposed system equips the end-to-end models with the streaming capability and reduces the large footprint from the streaming attention-based model using augmented memory \cite{wu2020streaming,shi2020weak}. 
On the {\it LibriSpeech} \cite{panayotov2015librispeech} dataset, our proposed system achieves word error rates $2.7\%$ on \texttt{test-clean} and $5.8\%$ on \texttt{test-other}, to our best knowledge the lowest among streaming approaches reported so far.

\end{abstract}


\begin{figure*}[hhh]
    \centering
    \includegraphics[scale=0.18]{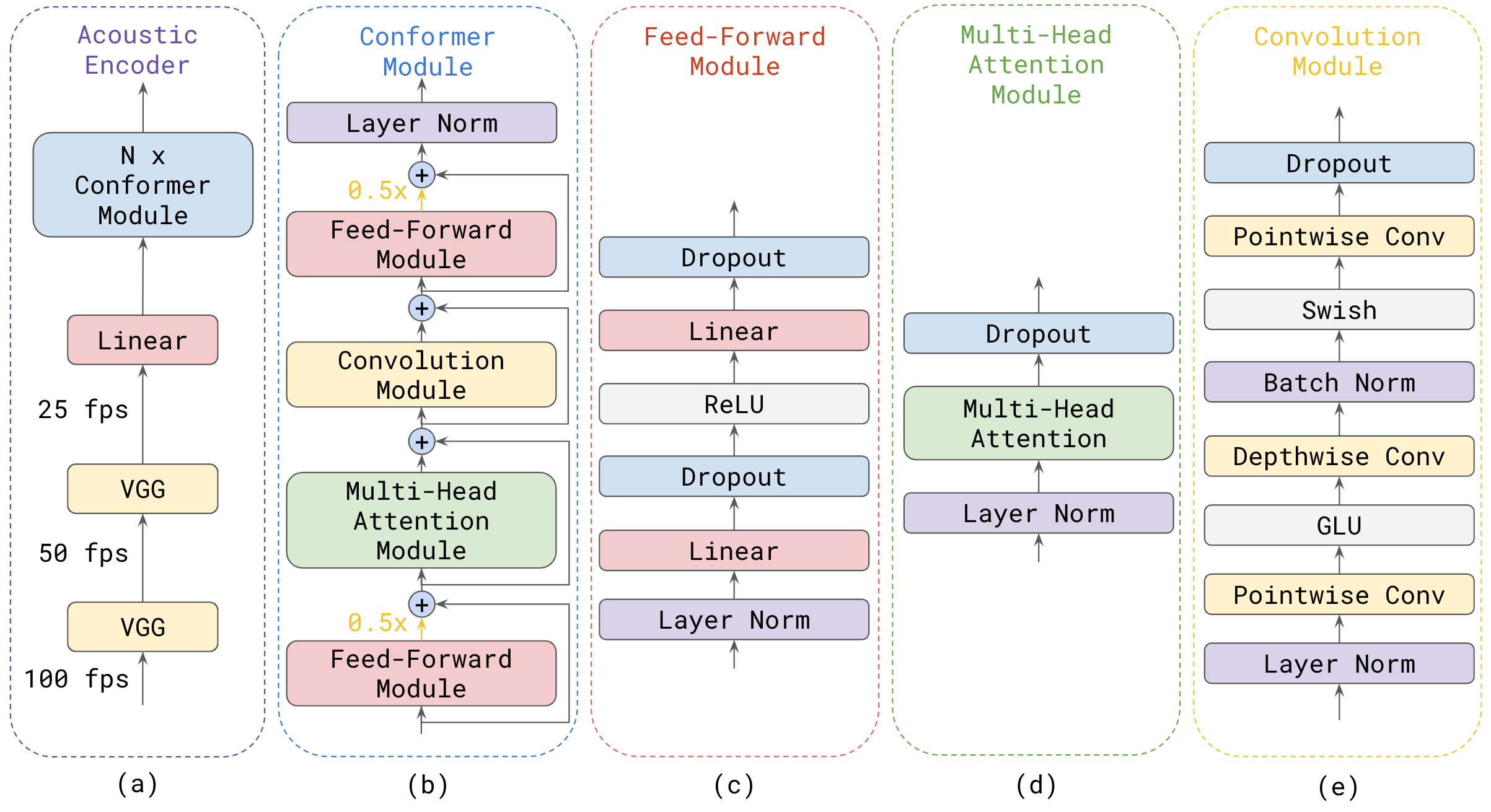}
    \caption{Decomposition of Conformer as Acoustic Encoder: (a) Acoustic Encoder (b) Conformer Module (c) Feed-forward Module (d) Multi-head Attention Module (e) Convolution Module.}
    \label{fig:model}
\end{figure*}

\section{Introduction}
\label{sec:introduction}
Sequence modeling is the core of speech recognition. 
For both more conventional ``hybrid" systems \cite{wang2019transformer, zhang2020fast} and more recently popular ``end-to-end" systems \cite{chiu2018state, he2019streaming}, neural encoders are used to extract high-level embeddings as the representation of input sequences.

Recurrent neural networks (RNNs) are naturally effective as encoders for speech signals given the recurrent connection between embeddings at different time steps.
Among variants of RNNs, long short-term memory (LSTM) \cite{hochreiter1997long} models are especially famous for its gated mechanism for capturing long and short term change.
However, the recurrent nature of RNNs also presents challenges.
For example, RNNs connects the previous state $\boldsymbol{h}_{t-1}$ with the current state $\boldsymbol{h}_t$ recurrently.
This way, the carried over information is condensed into fixed-sized vectors, and the direct connections between distant frames are limited, thus making capturing long contexts difficult.
Besides, the computation of $\boldsymbol{h}_t$ can only begin once the inference of $\boldsymbol{h}_{t-1}$ is finished; this makes the parallel computation of RNNs difficult.

In contrast to RNNs, attention-based models allow the computation to be fully parallel and explicitly connect the input tokens at any two positions directly, and have demonstrated significant success in fields such as machine translation \cite{vaswani2017attention} and speech recognition \cite{gulati2020conformer, wang2019transformer, zhang2020fast, zhang2020transformer}, and often referred to as ``Transformer" models.

Although attention-based models have achieved several milestones in sequence modeling, there remain challenges to be tackled with.
In addition to accuracy, latency is another important metric for quality evaluation of speech recognition systems.
For applications requiring low-latency, such as live captioning or messaging, the system often needs to be ``streaming", i.e., transcribing received segments as they arrive rather than waiting until the end of the utterance.
Streaming is a major challenge for attention-based models, as the extraction of the embedding at any position depends on all input tokens, or the access to the full input sequence is need for inference to begin.
Among various approaches proposed for enabling streaming \cite{wu2020streaming, zhang2020transformer, dai2019transformer, povey2018time, yeh2019transformer}, {\it attention with augmented memory} \cite{wu2020streaming} demonstrated several advantages including accuracy, parameter-efficiency and scalability. 
In addition, while being strong on long context modeling, attention-based models lack explicit modeling on local patterns, as shown with improvements from approaches such as positional encoding \cite{dai2019transformer} for NLP tasks.
However, speech signals pose additional challenges, such as being continuous rather than discrete and typically having longer sequence lengths (acoustic frames) than word tokens. 
Approaches including {\it weak-attention suppression} \cite{shi2020weak} and {\it convolution-augmented attention} \cite{gulati2020conformer} take these characteristics into consideration and further improve attention-based models for speech recognition.

Each of the mentioned related works has benefits and limitations on its own.
Specifically, {\it Conformer-Transducer} \cite{gulati2020conformer} is accurate and compact but streaming is a challenge, while {\it attention with augmented memory} \cite{wu2020streaming} and {\it weak-attention suppression} \cite{shi2020weak} were evaluated with hybrid systems which are commonly larger in footprint. 
In this work, we adopted the mentioned works to produce an integrated system that is accurate, compact, and streaming. 

\begin{figure*}[htbp]
    \centering
    \includegraphics[scale=0.2]{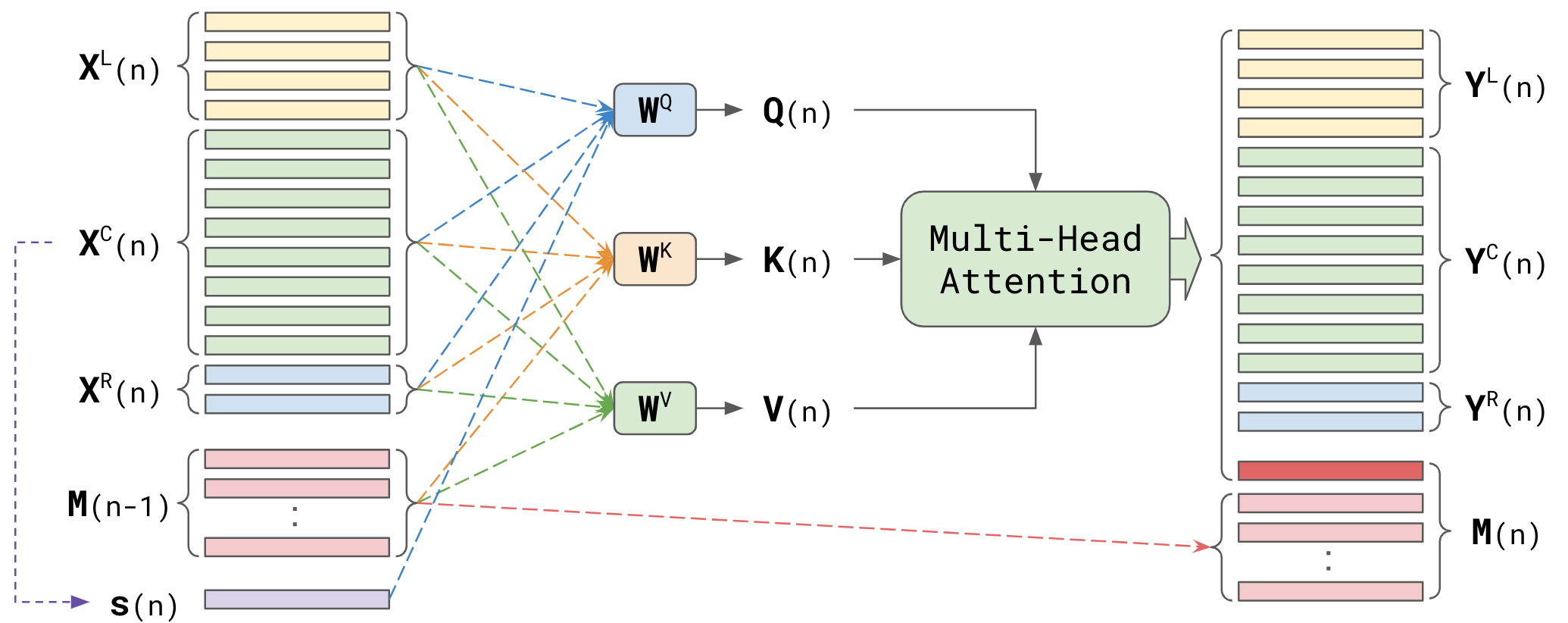}
    \caption{Streaming Attention with Augmented Memory}
    \label{fig:amtrf}
\end{figure*}

\section{Conformer-Transducer}
\label{sec:conformer}

Neural transducer (RNN-T) \cite{graves2012sequence} has demonstrated strong performance in the field of speech recognition and gained significant research interests \cite{gulati2020conformer, he2019streaming, zhang2020transformer, yeh2019transformer, han2020contextnet}. 
Compared with the traditional hybrid framework, neural transducer aims to model the transformation from speech signal to word tokens directly; therefore, the model becomes simpler, requires less human intervention, and is more compact in terms of system size.
Among many end-to-end approaches, such as Transformer \cite{vaswani2017attention, mohamed2019transformers} and Listen-Attend-and-Spell (LAS) \cite{chan2016listen}, the major benefit of the neural transducer is its streaming design, which is crucial for low-latency scenarios.

Inside neural transducer, the ``encoder" (similar to the ``acoustic model" in the hybrid framework) can be constructed with different neural network components, from the more traditional LSTM \cite{he2019streaming} to the recently popular attention-based modules \cite{wang2019transformer, zhang2020transformer, yeh2019transformer}. 
Attention-based models are strong and effective in long-term temporal modeling. They are widely adopted in recent works that produced new milestones in the field \cite{wang2019transformer, zhang2020fast, zhang2020transformer} and are often referred to as ``Transformer" models. 
While attention-based modules are strong on long-term modeling, they often suffer at localized and sequential patterns, which are particularly common given the acoustic frames in speech are highly correlated. 
Techniques such as positional encoding \cite{dai2019transformer} have addressed the issue, but the limitation remains. 
On the other hand, convolutional modules are naturally effective in extracting localized information, and recent works have shown improvements from the heterogeneous combination of attention-based modules and convolutional modules, i.e., the {\it Conformer} architecture \cite{gulati2020conformer}.
In addition, convolutional modules also demonstrated high parameter efficiency and a good tradeoff between accuracy and system size \cite{han2020contextnet, kriman2020quartznet}. 

Neural transducer with the encoder constructed with the Conformer architecture (or {\it Conformer-Transducer}) is accurate and effective for speech recognition.
Similar to other attention-based models, the major limitation of Conformer is the need of access to the full sequence in its attention modules, which blocks it from direct application to low-latency scenarios and not fully utilizing the streaming potential of the neural transducer.
In this work, we adopted Conformer-Transducer as the strong baseline system and applied (1) augmented memory (2) weak attention suppression to enable streaming while remaining competitive accuracy with non-streaming versions.

\begin{table}[htbp]
    \centering
    \caption{Transformer/Conformer Model Hyper-parameters.}
    \vspace{3mm}
    \scalebox{1.0}{
    \begin{tabular}{|c||cc|cc|}
    \hline
    \textbf{Attribute} & \multicolumn{2}{|c|}{\textbf{Transformer}} & \multicolumn{2}{|c|}{\textbf{Conformer}} \\
    \hline
    \hline
    \multicolumn{1}{|l||}{Code} & (S) & (M) & (S) & (M) \\
    \multicolumn{1}{|l||}{Parameters (M)} & 10.9 & 30.5 & 10.3 & 27.9  \\
    \hline
    \hline
    \multicolumn{1}{|l||}{Num. Layers} & 16 & 16 & 16 & 16 \\
    \multicolumn{1}{|l||}{Layer Dim.} & 160 & 288 & 144 & 256 \\
    \multicolumn{1}{|l||}{Attention Heads} & 4 & 4 & 4 & 4 \\
    \multicolumn{1}{|l||}{Conv. Kernels} & -- & -- & 32 & 32 \\
    \hline
    \end{tabular}
    }
    \label{tab:model_archs}
\end{table}

The model architecture of the encoder of the Conformer-Transducer in this work is illustrated in Fig. \ref{fig:model}. 
Fig. \ref{fig:model}(a) refers to the acoustic encoder component, where the acoustic frames first go through two VGG layers, each with subsampling factor 2 for reducing sequence length and adding inter-frame correlation information \cite{mohamed2019transformers}. 
Next, a linear layer applies projection on the embedding dimension before going to the following multiple Conformer modules.
The decomposition of the Conformer module is shown in Fig. \ref{fig:model}(b), where a multi-head attention module followed by a convolutional module forms the core component and are encapsulated by two macaron-like feed-forward modules, then followed by a post layer norm.
The feed-forward modules, multi-head attention modules and convolutional modules in Fig. \ref{fig:model}(b) are then further decomposed in Fig. \ref{fig:model}(c), Fig. \ref{fig:model}(d) and Fig. \ref{fig:model}(e) respectively, following the design in \cite{gulati2020conformer}. 
The hyper-parameters for models within several target size constraints are listed in Table. \ref{tab:model_archs}, where for Transformer modules, we compensate the input dimensions to match the size of corresponding Conformer modules for fair comparison.
We focused on compact models in this work and included only models with sizes around 10M and 30M, coded with (S) and (M), respectively.

\section{Attention with Augmented Memory} 
\label{sec:amtrf}

The main challenge of attention-based models to be streaming is the need for access to the full input sequence for inference.
To enable streaming, {\it attention with augmented memory} \cite{wu2020streaming} was proposed, tackling the challenge from two aspects.
First, for block-wise inference \cite{dong2019self}, the full input sequence is broken down into fix-sized segments with overlapping left context frames and right context frames.
By having the lengths of the segments fixed, the computational cost in attention modules becomes constant for individual segments and linear for the entire sequence, which also enables the processing of long sequences. 
Second, to carry over temporal dependency across segments, an {\it augmented memory bank} is used, where segment-wise information is summarized and stored in the form of embeddings in slots in the augmented memory bank.
The augmented memory bank demonstrated efficient and effective long-term modeling for speech signals \cite{wu2020streaming} compared with other streaming approaches for attention-based models such as Transformer XL \cite{dai2019transformer} and time-restricted self-attention \cite{povey2018time}.

Fig. \ref{fig:amtrf} illustrates how attention module works with augmented memory, with formulation in equations (\ref{eq:q}), (\ref{eq:k}), (\ref{eq:v}), (\ref{eq:m}) and (\ref{eq:M}).
Assuming the input sequence is broken down into $N$ segments of fixed length and padded with overlapping left/right contexts.
For the $n$-th segment, $\boldsymbol{X}^L(n)$, $\boldsymbol{X}^C(n)$, and $\boldsymbol{X}^R(n)$ represent the left context, the central body and the right context of the segment respectively, in the form of stacked embeddings, i.e., matrices.
A {\it summary embedding} $\boldsymbol{s}(n)$ is then computed by pooling over the central body $\boldsymbol{X}^C(n)$, providing the later multi-head attention additional information on segment-wise normalization.
We followed the original augmented memory work \cite{wu2020streaming} and used average pooling in this work.
\begin{align}
    \boldsymbol{Q}(n)&=\mathbf{W}^{\rm Q}[
        \boldsymbol{X}^L(n),
        \boldsymbol{X}^C(n),
        \boldsymbol{X}^R(n),
        \boldsymbol{s}(n)
    ], \label{eq:q} \\[3pt]
    \boldsymbol{K}(n)&=\mathbf{W}^{\rm K}[
        \boldsymbol{M}(n-1),
        \boldsymbol{X}^L(n),
        \boldsymbol{X}^C(n),
        \boldsymbol{X}^R(n)
    ], \label{eq:k} \\[3pt]
    \boldsymbol{V}(n)&=\mathbf{W}^{\rm V}[
        \boldsymbol{M}(n-1),
        \boldsymbol{X}^L(n),
        \boldsymbol{X}^C(n),
        \boldsymbol{X}^R(n)
    ], \label{eq:v} \\[3pt]
    \boldsymbol{m}(n)&={\textstyle\sum}_{i}\hspace{1mm}\Phi(\alpha_{n,i}) \cdot \boldsymbol{V}(n, i), \label{eq:m} \\[3pt]
    \boldsymbol{M}(n)&=[\boldsymbol{M}(n-1), \boldsymbol{m}(n)]. \label{eq:M}
\end{align}

The {\it query}, {\it key} and {\it value} for multi-head attention are then formed by these elements.
$\boldsymbol{X}^L(n)$, $\boldsymbol{X}^C(n)$, $\boldsymbol{X}^R(n)$ and $\boldsymbol{s}(n)$ are concatenated and go through projection $\boldsymbol{W}^Q$ to form the {\it query} $\boldsymbol{Q}(n)$. 
The concatenation of $\boldsymbol{M}(n-1)$, $\boldsymbol{X}^L(n)$, $\boldsymbol{X}^C(n)$ and $\boldsymbol{X}^R(n)$ goes through projections $\mathbf{W}^K$ and $\mathbf{W}^V$ separately to form the {\it key} $\boldsymbol{K}(n)$ and the {\it value} $\boldsymbol{V}(n)$. 
The formulation in equations (\ref{eq:q}), (\ref{eq:k}) and (\ref{eq:v}) makes $\boldsymbol{Q}(n)$ only contains information of current segment, while $\boldsymbol{K}(n)$ and $\boldsymbol{V}(n)$ contain carried over information in the augmented memory.
This allows the attention module to model the attention from the current segment on carried over history and generate output embeddings accordingly.
To carry over the information of the current segment to the next, the attention probabilities of {\it summary embedding} $\boldsymbol{s}(n)$ on the embeddings in the {\it value} $\boldsymbol{V}(n)$ are used as weights to aggregate the embeddings in $\boldsymbol{V}(n)$ to form the new memory slot $\boldsymbol{m}(n)$.
Here $\alpha_{n,i}$ is the attention probability of $\boldsymbol{s}(n)$ on the $i$-th embedding in $\boldsymbol{V}(n)$, and dropout is used as regularization function $\Phi$.
Finally, the new memory slot $\boldsymbol{m}(n)$ is added to the existing memory bank $\boldsymbol{M}(n-1)$ to form $\boldsymbol{M}(n)$, to be used for the next segment.
Note that the output embeddings generated also contain the left context $\boldsymbol{Y}^L(n)$, the central body $\boldsymbol{Y}^C(n)$ and the right context $\boldsymbol{Y}^R(n)$.
They are propagated through attention layers and only stripped out at the output of the acoustic encoder.
This guarantees the contexts stay constant with a growing number of layers, which makes the approach scalable and the latency controllable.

\section{Weak-Attention Suppression (WAS)}
\label{sec:was}
The attention module models the distribution of attention probabilities between embeddings in parallel, enabling better long-term temporal modeling and additional mechanisms to handle ambiguity between embeddings at different positions, such as positional encodings \cite{dai2019transformer}. 
Different from NLP tasks, speech signals come in continuous forms and typically are significantly longer.
For example, a 3-second long utterance may contain 10 word tokens but 300 acoustic frames.
This increases the difficulty of attention modeling due to the high correlation between acoustic frames \cite{shi2020weak}, as the attention probability may be diluted among similar embeddings while a localized and sparse distribution can be more robust.
Weak-attention suppression (WAS) \cite{shi2020weak} was proposed to improve the attention probability distribution by forcing the model to skip the embeddings with low attention probabilities dynamically.

The suppression is a two-stage process: (1) threshold determination, (2) attention weight modification.
For threshold determination, a dynamic threshold ${\theta}_i$ is selected based on the attention distribution for the $i$-th position of {\it query} $\boldsymbol{Q}$:
\begin{align}
    {\theta}_i  = {\mu}_i - \gamma \cdot {\sigma}_i. \label{eq:theta}
\end{align}
where ${\mu}_i$ and ${\sigma}_i$ are the mean and standard deviation of the attention probabilities at the $i$-th position of $\boldsymbol{Q}$, and $\gamma$ is a user-specified hyper-parameter to control the level of suppression.
For attention weight modification, the attention weights ${\omega}_{i,j}$ from positions $i$ to $j$ corresponding to the indices having probability lower than ${\theta}_i$ are set to negative infinity:
\begin{align}
    \hat{\omega}_{i,j}  = \begin{cases}
      {\omega}_{i,j}, & \text{if } p_{i,j} \geq {\theta}_i \\
      -{\infty},              & \text{otherwise}
    \end{cases}.
    \label{eq:omega}
\end{align}
where $p_{i,j}$ and ${\omega}_{i,j}$ are the attention probability and the attention weight from positions $i$ to $j$.
With modified attention weights, the produced attention probabilities for suppressed positions will be 0 while the entire distribution will be re-normalized.

\begin{table*}[htbp]
    \centering
    \caption{Comparison on Systems and Architectures on {\it LibriSpeech} \cite{panayotov2015librispeech} (Non-streaming).}
    \vspace{3mm}
    \scalebox{1.0}{
    \begin{tabular}{|c|c|c||cc|cc|}
    \hline
    \textbf{System} & \textbf{Model} & \textbf{Size (M)} &
        \multicolumn{2}{|c|}{\begin{tabular}{@{}c@{}}\textbf{WER w/o LM} \\ \textbf{test-\{clean,other\}}\end{tabular}} &
        \multicolumn{2}{|c|}{\begin{tabular}{@{}c@{}}\textbf{WER w/ LM} \\ \textbf{test-\{clean,other\}}\end{tabular}} \\
    \hline
    \hline
    \multirow{2}{*}{Hybrid}
        & \multicolumn{1}{|l|}{Transformer \cite{wang2019transformer}} & $\simeq$80 
            & \parbox{1.2cm}{\centering--} & \parbox{1.2cm}{\centering--} & \parbox{1.2cm}{\centering2.3} & \parbox{1.2cm}{\centering4.9} \\
        & \multicolumn{1}{|l|}{Transformer \cite{zhang2020fast}} & $\simeq$124 & -- & -- & \textbf{2.1} & \textbf{4.2} \\ 
    \hline
    \hline
    \multirow{4}{*}{Transducer} 
        & \multicolumn{1}{|l|}{Transformer \cite{zhang2020transformer}} & $\simeq$139 & 2.4 & 5.6 & 2.0 & 4.6 \\ 
        & \multicolumn{1}{|l|}{Conformer (S) \cite{gulati2020conformer}} & 10.3 & 2.7 & 6.3 & 2.1 & 5.0 \\ 
        & \multicolumn{1}{|l|}{Conformer (M) \cite{gulati2020conformer}} & 30.7 & 2.3 & 5.0 & 2.0 & 4.3 \\ 
        & \multicolumn{1}{|l|}{Conformer (L) \cite{gulati2020conformer}} & 118.8 & \textbf{2.1} & \textbf{4.3} & \textbf{1.9} & \textbf{3.9} \\ 
    \hline
    \hline
    \multirow{4}{*}{\begin{tabular}{@{}c@{}}Transducer \\ (Our Re-implementation)\end{tabular}}
        & \multicolumn{1}{|l|}{Transformer (S)} & 10.9 & 14.4 & 18.5 & 13.2 & 16.2 \\ 
        & \multicolumn{1}{|l|}{Transformer (M)} & 30.5 & 9.4  & 11.4 & 8.8 & 10.3 \\ 
        & \multicolumn{1}{|l|}{Conformer (S)}   & 10.3 & 3.0 & 6.8 & 2.4 & 5.4 \\ 
        & \multicolumn{1}{|l|}{Conformer (M)}   & 27.9 & \textbf{2.5} & \textbf{5.5} & \textbf{2.2} & \textbf{4.7} \\ 
    \hline
    \end{tabular}
    }
    \label{tab:libri_offline}
\end{table*}

\section{Experiments}
\label{sec:experiments}

The models were trained and evaluated using an in-house extension of the PyTorch-based \emph{fairseq}~\cite{ott2019fairseq} toolkit.
In all experiments, 80-dimensional log Mel-filter bank features with a 10ms frame-shift and a 25ms frame-width were used as input features.
Speed perturbation and SpecAugment \cite{park2019specaugment} were applied to enhance the robustness and overall accuracy of the systems.

For transducer systems, we built sentence piece models \cite{kudo2018sentencepiece} with 1024 targets across experiments to ensure identical system sizes.
Byte pair encoding (BPE) \cite{Sennrich_2016} was adopted as the segmentation algorithm.
We focused on identifying the impact of the encoder (or the ``acoustic encoder") in neural transducer and used identical architectures for the predictor (or the ``label encoder") and the joiner (or the ``joint network").
For the predictor, the tokens are first represented by 256-dimensional embeddings before going through a single LSTM layer with 320 hidden nodes, followed by a linear projection to 640-dimensional features before the joiner.
For the joiner, the combined embeddings from the encoder and the predictor first go through a tanh activation and then another linear projection to the target number of sentence pieces (1024).
Relative positional encoding \cite{dai2019transformer} was applied in both Transformer and Conformer models in experiments.
For models with augmented memory, we applied 16 frames as the left context, 32 frames as the central body, and 8 frames as the right context, which translates to a 320ms right context (excluding computational delay) in block-wise processing given the subsampling factor 4 from the VGG layers. 
We applied a universal $\gamma = 0.5$ for weak-attention suppression (WAS) for all models involved. 

\vspace{-1mm}
\subsection{LibriSpeech: Description and Setup}
The {\it LibriSpeech} dataset \cite{panayotov2015librispeech} contains about 960 hours of reading speech data for training and an additional 800M word tokens text-only corpus for building language models. 
The word error rates on sets \texttt{test-\{clean,other\}} are reported.
For hybrid systems, the standard 4-gram language model with a 200K vocabulary was used for all first-pass decoding, which contains roughly 144M n-gram probabilities, excluding the backoff weights.
For transducer systems, the transcript of the \texttt{train} set was combined with the 800M text-only set to build LSTM-based neural network language models (NNLMs), similar to \cite{gulati2020conformer}.
We experimented with the NNLMs with a fixed number of layers but different hidden nodes per layer (3x512, 3x1024, 3x2048, and 3x4096) to evaluate the tradeoff between WER and parameter efficiency, while the largest model (3x4096) is used for demonstrating the best case WERs for individual systems. 
We interpolate NNLM probabilities with the raw probabilities of the hypotheses with constant weight 0.25 during beam search in the on-the-fly manner, referred to as ``shallow fusion" \cite{gulcehre2015using}, which is streaming by itself and applicable to both streaming and non-streaming models.

\subsection{Results on System and Architecture}
Table \ref{tab:libri_offline} presents the comparison on model framework and architecture.
All attention modules in the models are non-streaming, i.e., they need access to the full sequence for inference. 
Note that the system size for hybrid systems here includes only the acoustic model and excludes both the n-gram LM and the NNLM, while the ``transducer" system size includes all components (encoder, predictor, and joiner \cite{yeh2019transformer}).
With attention modules as the core of the acoustic model, hybrid systems achieved impressive accuracy along with both n-gram and neural language models on LibriSpeech, with the cost of high-latency (non-streaming).
From the results of \cite{zhang2020transformer}, transducer-based models proved to be similarly strong on WERs but able to improve the parameter efficiency by removing external language models (both the n-gram and neural ones).
For example, the hybrid system in \cite{zhang2020fast} gives 2.1 and 4.2 on \texttt{test-\{clean,other\}} with a $\simeq$124M acoustic model, a $\simeq$144M n-gram LM and a $\simeq$351M neural LM, while the transducer system in \cite{zhang2020transformer} gives 2.4 and 5.6 with a $\simeq$139M model alone.
This demonstrates that parameter efficiency (or the overall system size) is one of the major advantages of transducer systems.

In addition, comparing the Conformer models in \cite{gulati2020conformer} with the Transformer models in \cite{zhang2020transformer}, we see that the introduction of convolutional modules is significantly effective both in terms of accuracy and parameter efficiency, from the 30.7M Conformer and the $\simeq$139M Transformer results. 
Although we could not fully reproduce the results in \cite{gulati2020conformer}, we observed similar trends by comparing Transformer and Conformer models on similar scales from our re-implementation, listed as ``Transducer (Our Re-implementation)". 
The Transformer models for comparison here follows the same architecture as the Conformer models illustrated in Fig. \ref{fig:model}, with convolutional modules removed and dimension increased to compensate for the model sizes as listed in Table. \ref{tab:model_archs}.
It is evident that within the same range of model size, the Conformer models significantly outperform the Transformer models, demonstrating the heterogeneous combination of attention-based and convolutional modules.

\begin{table*}[htbp]
    \centering
    \caption{Comparison on Streaming and Non-streaming Models on {\it LibriSpeech} \cite{panayotov2015librispeech}.}
    \vspace{3mm}
    \scalebox{1.0}{
    \begin{tabular}{|c|c|c|c||cc|}
    \hline
    \textbf{System} & \textbf{Model} & \textbf{Size (M)} & \textbf{Streaming} & 
        \multicolumn{2}{|c|}{\begin{tabular}{@{}c@{}}\textbf{WER} \\ \textbf{test-\{clean,other\}}\end{tabular}} \\
    \hline
    \hline
    \multirow{3}{*}{Hybrid}
        & \multicolumn{1}{|l|}{Transformer \cite{zhang2020fast} (+NNLM)} & \multirow{1}{*}{$\simeq$124} & \multirow{2}{*}{No} & \parbox{1.2cm}{\centering2.1} & \parbox{1.2cm}{\centering4.2} \\ \cline{3-2}
        & \multicolumn{1}{|l|}{Transformer \cite{wu2020streaming}} & \multirow{2}{*}{$\simeq$80} & & 2.6 & 5.6 \\ \cline{4-4}
        & \multicolumn{1}{|l|}{Transformer \cite{wu2020streaming} + Aug. Mem.} & & Yes & \textbf{2.8} & \textbf{6.7} \\
    \hline
    \hline
    \multirow{2}{*}{Transducer} 
        & \multicolumn{1}{|l|}{Conformer (S) \cite{gulati2020conformer} (+NNLM)} & 10.3 & \multirow{2}{*}{No} & 2.1 & 5.0 \\ 
        & \multicolumn{1}{|l|}{Conformer (M) \cite{gulati2020conformer} (+NNLM)} & 30.7 & & 2.0 & 4.3 \\ 
    \hline
    \hline
    \multirow{10}{*}{\begin{tabular}{@{}c@{}}Transducer \\ (Our Re-implementation)\end{tabular}}
        & \multicolumn{1}{|l|}{Conformer (S)}   & \multirow{5}{*}{10.3} & \multirow{2}{*}{No} & 3.0 & 6.8 \\ 
        & \multicolumn{1}{|l|}{\hspace{3mm}+ NNLM} &  &  & \textbf{2.4} & \textbf{5.4} \\ \cline{4-4}
        & \multicolumn{1}{|l|}{Conformer (S) + Aug. Mem.} &  & \multirow{3}{*}{Yes} & 3.6 & 8.3 \\ 
        & \multicolumn{1}{|l|}{\hspace{3mm}+ WAS} &  &  & 3.4 & 8.0 \\ 
        & \multicolumn{1}{|l|}{\hspace{3mm}+ NNLM} &  &  & \textbf{3.1} & \textbf{6.6} \\ \cline{3-4}
        & \multicolumn{1}{|l|}{Conformer (M)}   & \multirow{5}{*}{27.9} & \multirow{2}{*}{No} & 2.5 & 5.5 \\ 
        & \multicolumn{1}{|l|}{\hspace{3mm}+ NNLM} &  &  & \textbf{2.2} & \textbf{4.7} \\ \cline{4-4}
        & \multicolumn{1}{|l|}{Conformer (M) + Aug. Mem.} &  & \multirow{3}{*}{Yes} & 3.1 & 6.5 \\ 
        & \multicolumn{1}{|l|}{\hspace{3mm}+ WAS} &  &  & 3.0 & 6.4 \\ 
        & \multicolumn{1}{|l|}{\hspace{3mm}+ NNLM} &  &  & \textbf{2.7} & \textbf{5.8} \\
    \hline
    \end{tabular}
    }
    \label{tab:libri_streaming}
\end{table*}

\subsection{Results on Streaming Models}
Table \ref{tab:libri_streaming} presents the comparison between streaming and non-streaming models.
For hybrid systems, introducing augmented memory to attention modules enables streaming and improves latency with the cost of increased WERs. 
Similar trade-offs were observed for transducer systems, for example, the degradation from 2.5/5.5 to 3.1/6.5 with the 27.9M Conformer model.
We further applied weak-attention suppression (WAS) on top of the results with augmented memory.
Although the magnitude of relative improvement observed was less significant compared with hybrid system results \cite{wu2020streaming, shi2020weak}, possibly from overlapping functionalities by convolutional modules in Conformer, weak-attention showed consistent improvements and reduced the gap from non-streaming models. 
On top of that, shallow fusion with NNLM further improves the results by the streaming hybrid system \cite{wu2020streaming} from 2.8/6.7 to 2.7/5.8. 
In this comparison, both the hybrid and the transducer systems benefit from the 800M text-only corpus, and the language models involved are applied in streaming fashions during beam search, therefore both remain low-latency.

\subsection{Results on Shallow Fusion with NNLMs}

\begin{figure}[htbp]
    \centering
    \includegraphics[scale=0.13]{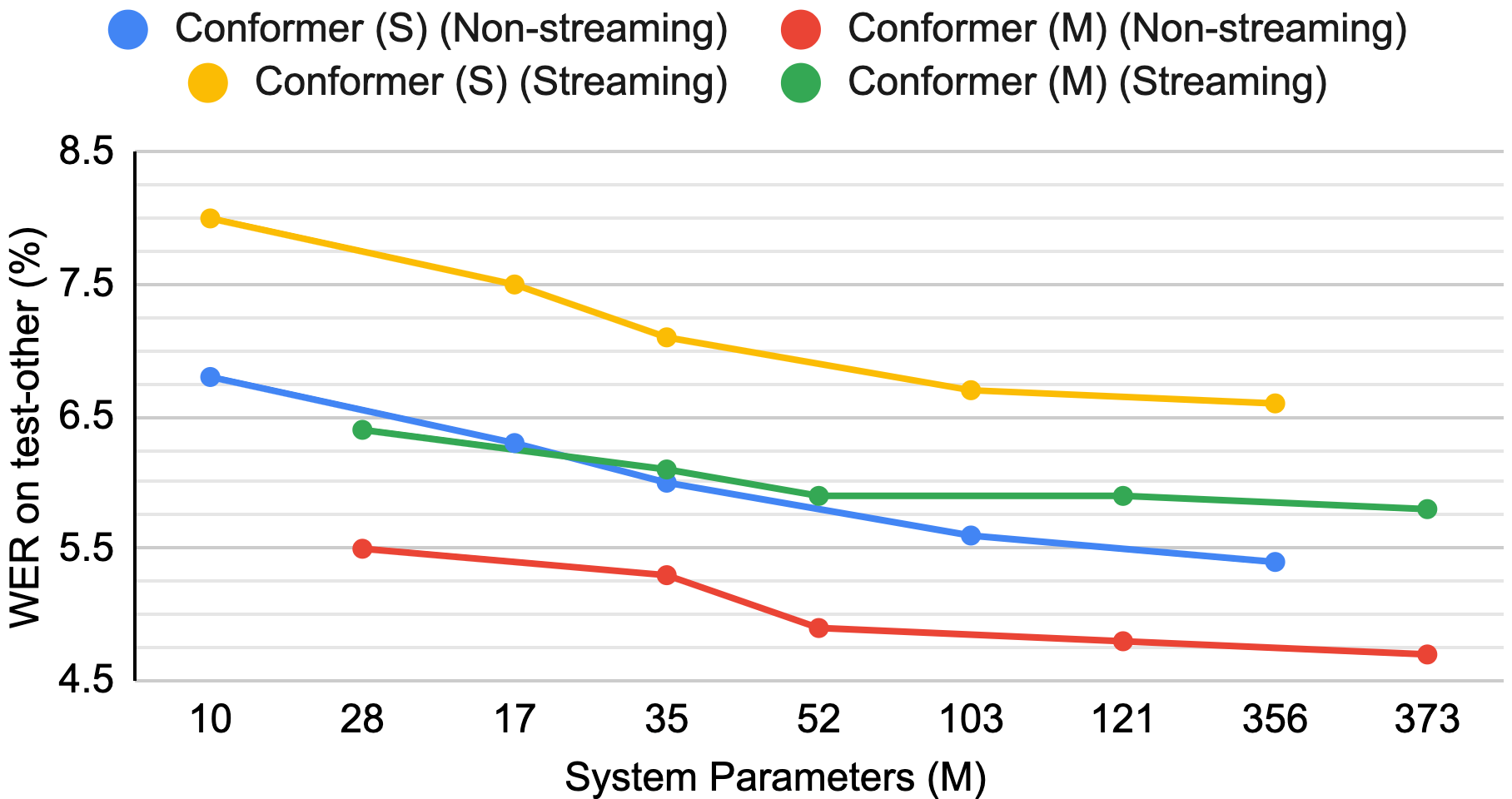}
    \caption{Comparison on NNLMs of Different Sizes.}
    \label{fig:nnlm}
\end{figure}

From Table \ref{tab:libri_streaming}, it is clear that an external NNLM consistently helps on top of transducer models alone.
However, NNLMs also contribute significantly to the overall system size.
For example, the NNLM used in Table \ref{tab:libri_streaming} has 3 LSTM layers with 4096 hidden nodes each, translating to $\simeq$345.3M parameters, which is quite disproportional to the size of transducer models (10.3M or 27.9M).
For further investigation, we evaluated NNLMs with different sizes and plotted results in Fig. \ref{fig:nnlm} to demonstrate the trade-off between WERs and system sizes.
The number of LSTM layers in the NNLM was kept the same, while the number of hidden nodes varies.
The architectures evaluated include 3x512 ($\simeq$ 6.8M), 3x1024 ($\simeq$24.4M), 3x2048 ($\simeq$92.6M) and 3x4096 ($\simeq$345.3M).
WERs against the combined system sizes of model and NNLM are plotted to show the parameter efficiency across models and NNLMs.
We see that all NNLMs of different sizes help, but the effectiveness does not grow linearly with the number of parameters.
Since the computational cost and latency grow with the number of parameters in practice, this indicates that in pursuit of compact and low-latency systems, trade-offs between word error rates and size of NNLMs may be worth further investigations.

From Fig. \ref{fig:nnlm}, we see that the improvement from NNLMs saturates more quickly for the larger 27.9M Conformer models, both streaming and non-streaming, compared with the smaller 10.3M ones.
This shows that it is more efficient to distribute more parameters on the model than on the NNLM, given constant budgets on overall system sizes. 
In addition, from the comparison between streaming and non-streaming trends, the improvement from NNLMs appears consistent and is not entangled with whether the model is streaming or not.
From the trade-offs shown in Fig. \ref{fig:nnlm}, a 27.9M Conformer model with a 3x1024 ($\simeq$24.4M) NNLM ($\simeq$52.3M combined) can achieve similar word error rate with less than 5\% relative degradation compared with the results with a 3x4096 ($\simeq$345.3M) NNLM, but only $\simeq$15\% in size.

\section{Conclusion}
\label{sec:conclusion}

In this work, we investigated several cutting-edge technologies in speech recognition and analyzed each one's strengths and challenges. 
The {\it Conformer-Transducer} offers high accuracy, but streaming remains to be challenging. 
With the {\it augmented memory}, the model propagates the contextual dependency from previous segments in attention modules, which enables streaming.
The {\it weak-attention suppression} focuses on localized pattern modeling to improve attention modules.
By integrating all three technologies, they contribute complementary benefits to compensate for the limitations of each.
Specifically, the transducer framework offers a compact system size, convolution-augmented attention module and weak attention suppression ensure high accuracy, augmented memory unblocks streaming and reduces latency.
On the \texttt{test-\{clean,other\}} sets of the widely used {\it LibriSpeech} dataset, with the budget of 320ms lookahead, we observed word error rates of 3.0/6.4 for the system of 27.9M parameters alone, and 2.7/5.8 for the same system with an external neural language model, both of which are the lowest error rates among systems within a similar range of latency and sizes to our knowledge so far.
The integrated system demonstrated several major benefits, including high accuracy, high parameter-efficiency, and low-latency.

{
\clearpage
\small
\bibliographystyle{IEEEbib}
\bibliography{strings,refs}
}

\end{document}